\def\year{2021}\relax
\documentclass[letterpaper]{article} 
\usepackage{aaai21}  
\usepackage{times}  
\usepackage{helvet} 
\usepackage{courier}  
\usepackage[hyphens]{url}  
\usepackage{graphicx} 
\usepackage{natbib}  
\usepackage{caption} 

\usepackage{amsmath}
\usepackage{amsfonts}
\usepackage{amssymb}
\usepackage{amsthm}
\usepackage{booktabs}
\usepackage{caption}
\usepackage{changepage}
\usepackage{chngcntr}
\usepackage{comment}
\usepackage{csquotes}
\usepackage{draftcopy}
\usepackage{graphicx}
\usepackage{makecell}
\usepackage{mathtools}
\usepackage{microtype}
\usepackage{natbib}
\usepackage{sectsty}
\usepackage{subcaption}
\usepackage{tcolorbox}
\usepackage{thm-restate}
\usepackage{thmtools}
\usepackage{threeparttable}
\usepackage{url}
\usepackage{verbatim}

\usepackage[hidelinks]{hyperref}
\usepackage[capitalise,nameinlink]{cleveref}
\hypersetup{nolinks=true}

\def\<{\langle}
\def\>{\rangle}

\DeclarePairedDelimiter\abs{\lvert}{\rvert}
\makeatletter
\let\oldabs\abs
\def\abs{\@ifstar{\oldabs}{\oldabs*}}
\makeatother

\crefname{table}{Table}{Tables}
\crefname{figure}{Figure}{Figures}
\crefname{equation}{Equation}{Equations}
\creflabelformat{equation}{#2\textup{#1}#3}

\def\L{{\mathcal{L}}}
\def\hL{{\hat{\mathcal{L}}}}
\def\grL{{\nabla \mathcal{L}}}
\def\grhL{{\nabla \hat{\mathcal{L}}}}
\newcommand{\ie}{\textit{i}.\textit{e}.}
\newcommand{\eg}{\textit{e}.\textit{g}.}

\usepackage{amsmath,amsfonts,bm}

\def\eqref#1{equation~\ref{#1}}

\def\1{\bm{1}}

\def\eps{{\epsilon}}

\DeclareMathAlphabet{\mathsfit}{\encodingdefault}{\sfdefault}{m}{sl}
\SetMathAlphabet{\mathsfit}{bold}{\encodingdefault}{\sfdefault}{bx}{n}

\newcommand{\R}{\mathbb{R}}

\newcommand{\Var}{\mathrm{Var}}

\sectionfont{\centering}

\newcommand\blfootnote[1]{%
  \begingroup
  \renewcommand\thefootnote{}\footnote{#1}%
  \addtocounter{footnote}{-1}%
  \endgroup
}

\title{On the Adequacy of Untuned Warmup for Adaptive Optimization}

\author {
    Jerry Ma, \textsuperscript{\rm 1~2}
    Denis Yarats \textsuperscript{\rm 3~4} \\
}
\affiliations {
    \textsuperscript{\rm 1} Booth School of Business, University of Chicago \\
    \textsuperscript{\rm 2} U.S. Patent and Trademark Office, Department of Commerce \\
    \textsuperscript{\rm 3} Courant Institute of Mathematical Sciences, New York University \\
    \textsuperscript{\rm 4} Facebook AI Research \\
    research@jerryma.net, denisyarats@cs.nyu.edu
}

\begin{document}

\maketitle

{
  \begin{abstract}
    Adaptive optimization algorithms such as Adam are widely used in deep learning. The stability of such algorithms is often improved with a warmup schedule for the learning rate. Motivated by the difficulty of choosing and tuning warmup schedules, recent work proposes automatic variance rectification of Adam's adaptive learning rate, claiming that this rectified approach (``RAdam'') surpasses the vanilla Adam algorithm and reduces the need for expensive tuning of Adam with warmup. In this work, we refute this analysis and provide an alternative explanation for the necessity of warmup based on the magnitude of the \emph{update term}, which is of greater relevance to training stability. We then provide some ``rule-of-thumb'' warmup schedules, and we demonstrate that simple untuned warmup of Adam performs more-or-less identically to RAdam in typical practical settings. We conclude by suggesting that practitioners stick to linear warmup with Adam, with a sensible default being linear warmup over $2 / (1 - \beta_2)$ training iterations.\blfootnote{Appendix available at \url{https://arxiv.org/abs/1910.04209}.}
  \end{abstract}

  \section{Introduction}
\label{sec:intro}

Stochastic gradient-based optimization serves as the workhorse training approach for many classes of parametric models, including neural networks. Stochastic gradient descent and its various first-order cousins~\citep{polyak1964momentum, nesterov1983nag} have enabled numerous advances in deep learning across domains~\citep{krizhevsky2012alexnet, he2016resnet, gehring2017convs2s}. More recently, adaptive optimization algorithms have become prevalent in training the largest deep learning models. These adaptive methods, which include Adagrad~\citep{duchi2010adagrad}, RMSProp~\citep{hinton2012rmsprop}, and Adam~\citep{kingma2014adam}, scale the step size for each individual parameter based on various gradient moments.

Many practitioners have adopted the Adam algorithm for general-purpose use; notably, the preponderance of recent state-of-the-art results in natural language processing~\citep{devlin2018bert, radford2019gpt2, liu2019roberta, brown2020gpt3} have employed Adam, demonstrating the algorithm's ability to effectively train neural networks with parameter counts from 100 million to several billion.

In these large-scale settings, Adam's global learning rate is usually annealed with a ``warmup schedule'' which promotes early-stage training stability by regulating the size of the parameter updates. The prevalent warmup schedule is a simple linear warmup, in which the global learning rate starts at zero and increases by a constant at each iteration until reaching its intended value.~\footnote{Linear warmup has also been deployed for first-order optimization -- see, for example, \citet{goyal2017imagenet}.} The parameters of these warmup schedules are typically tuned for each problem setting and model.

\citet{liu2019radam} performed an analysis of Adam with warmup, concluding that Adam requires a warmup schedule to mitigate the large or divergent variance of the per-parameter scale term. They then propose the rectified Adam (``RAdam'') algorithm, which automatically corrects for this high variance. \citeauthor{liu2019radam} highlight the robustness of RAdam, noting in particular that RAdam reduces or eliminates the need for tuning warmup schedules when using Adam. RAdam has been applied to domains including generative modeling~\citep{yamamoto2020parallel}, natural language processing~\citep{nguyen2019trans}, and video retrieval~\citep{liu2019video}.

\subsection*{Contributions}

Our contributions in this work are as follows:

\textbf{Reexamining RAdam and the variance-based motivation for warmup}
\quad
We dive into the inner operation of RAdam and find that it is precisely Adam with a fixed warmup schedule, with the only deviation being to perform four iterations of heavy-ball momentum~\citep{polyak1964momentum} at the outset. We then argue that the variance-based motivation for warmup is impaired as it overlooks the correlation between the first and second moment estimators, which is crucial for understanding the actual parameter updates applied by Adam.

\textbf{Analyzing Adam's early-stage update magnitudes}
\quad
Shifting focus from gradients to parameter updates, we then perform a simulation-based analysis on the magnitudes of Adam's parameter updates. We find that even at a simulated local minimum of the objective, Adam exhibits considerable non-regularity in its early-stage parameter updates, shedding light on why Adam may require learning rate warmup to a greater extent than first-order optimization methods.

\textbf{Demonstrating the sufficiency of untuned warmup}
\quad
We provide some simple and intuitive ``rule-of-thumb'' warmup schedules for Adam, all of which require no tuning. As our main empirical result, we demonstrate that these schedules result in substantively \emph{identical} performance and training dynamics to those of RAdam across a wide range of models, problem settings, and hyperparameters, indicating that any claimed benefits can be achieved with lower complexity using off-the-shelf optimization tools. As a sensible untuned default, we recommend linear warmup over $2 \cdot \left( 1 - \beta_2 \right)^{-1}$ iterations.
  \section{Preliminaries}
\label{sec:prelim}

We begin with notation and a brief review of stochastic gradient descent and Adam.

\textbf{Primitives} \quad $\theta \in \R^p$ denotes a vector of model parameters. $\L(\theta) : \R^p \rightarrow \R$ denotes a loss function to be minimized over the model parameters. $\hL(\theta) : \R^p \rightarrow \R$ denotes an unbiased approximator of the loss function (\eg{} over a minibatch). $\grL(\theta)$ and $\grhL(\theta)$ denote the gradient of $\L(\theta)$ and $\hL(\theta)$, respectively. The terms $\theta$, $\hL(\theta)$, and $\grhL(\theta)$ are subscriptable by $t \geq 0$, the optimization time step (``training iteration''). $\theta_0$ represents the initial model parameters.

We write optimization algorithms as per-iteration procedures (``update rules''), taking the basic form:
\begin{equation*}
    \theta_t \leftarrow \theta_{t - 1} - \underbrace{\left\{~~~~\ldots~~~~\right\}}_{\text{``update step''}}
\end{equation*}

\textbf{Stochastic gradient descent} \quad The SGD algorithm, parameterized by learning rate $\alpha > 0$, performs the following procedure at each iteration $t$:
\begin{equation}
    \theta_t \leftarrow \theta_{t - 1} - \alpha \cdot \grhL_{t - 1}(\theta_{t - 1}) \label{eqn:sgd-update-rule}
\end{equation}

\textbf{Adam} \quad The Adam algorithm~\citep{kingma2014adam}, parameterized by global learning rate $\alpha > 0$, discount factors $\beta_1, \beta_2 \in (0, 1)$, and stability constant $\epsilon > 0$, performs the following procedure at each iteration $t$:
\begin{align}
    m_t &\leftarrow \beta_1 \cdot m_{t - 1} + (1 - \beta_1) \cdot \grhL_{t - 1}(\theta_{t - 1}) \label{eqn:adam-first-moment-update} \\
    v_t &\leftarrow \beta_2 \cdot v_{t - 1} + (1 - \beta_2) \cdot \left[ \grhL_{t - 1}(\theta_{t - 1}) \right]^2 \label{eqn:adam-second-moment-update} \\
    \theta_t &\leftarrow \theta_{t - 1} - \alpha \left[ \frac{(1 - \beta_1^t)^{-1} \cdot m_t}{\sqrt{(1 - \beta_2^t)^{-1} \cdot v_t} + \epsilon} \right] \label{eqn:adam-update-rule}
\end{align}
where $m, v \in \R^p$ denote auxiliary memory (interpretable as first moment and second moment estimators of $\grhL_t$, respectively). By convention, $m_0 = v_0 = 0$.

\textbf{Warmup schedules} \quad For any optimization algorithm parameterized with a learning rate $\alpha$, a \emph{warmup schedule} $\omega$ can be applied. $\omega$ is a sequence of ``warmup factors'' $\omega_t \in [0, 1]$, which serve to dampen the step size of each iteration $t$. Specifically, a warmup schedule is imposed by replacing $\alpha$ with $\alpha_t = \alpha \cdot \omega_t$ in the algorithm's update rule.

Perhaps the most common functional form for the schedule is \emph{linear warmup}, parameterized by a ``warmup period'' $\tau$:
\begin{equation}
    \omega_t^{\text{linear}, \tau} = \min \left\{ 1, \frac{1}{\tau} \cdot t \right\}
\end{equation}

\textbf{Rectified Adam} \quad The RAdam algorithm~\citep{liu2019radam}, parameterized identically to Adam, performs the following procedure at each iteration $t$:
\begin{align}
    \rho_\infty &\leftarrow 2 / (1 - \beta_2) - 1 \label{eqn:radam-rhoinf} \\
    \rho_t &\leftarrow \rho_\infty - 2 t \cdot \beta_2^t / (1 - \beta_2^t) \label{eqn:radam-rhot} \\
    \omega_t &\leftarrow \sqrt{\frac{(\rho_t - 4) (\rho_t - 2) \rho_\infty}{(\rho_\infty - 4) (\rho_\infty - 2) \rho_t}} \label{eqn:radam-warmup} \\
    m_t &\leftarrow \beta_1 \cdot m_{t - 1} + (1 - \beta_1) \cdot \grhL_{t - 1}(\theta_{t - 1}) \label{eqn:radam-first-moment-update} \\
    v_t &\leftarrow \beta_2 \cdot v_{t - 1} + (1 - \beta_2) \cdot \left[ \grhL_{t - 1}(\theta_{t - 1}) \right]^2 \label{eqn:radam-second-moment-update} \\
    \theta_t &\leftarrow \theta_t - \begin{cases}
      \alpha \cdot (1 - \beta_1^t)^{-1} \cdot m_t & \rho_t \leq 4 \\
      \alpha \cdot \omega_t \cdot \left[ \frac{(1 - \beta_1^t)^{-1} \cdot m_t}{\sqrt{(1 - \beta_2^t)^{-1} \cdot v_t} + \epsilon} \right] & \rho_t > 4
    \end{cases} \label{eqn:radam-update-rule}
\end{align}
  \section{Rectified Adam, Adaptive Variance, and Update Steps}
\label{sec:radam}

\begin{figure*}[ht]
  \input{fig/emnist.tex}
\end{figure*}

We begin by uncovering the precise behavior of RAdam, before delving into its underlying variance-based motivation.

\subsection{RAdam: Perform 4 Iterations of Momentum SGD, Then Use Adam with Fixed Warmup}

\citeauthor{liu2019radam} describe RAdam as having two modes of operation: ``divergent variance'' and ``convergent variance'', corresponding respectively to the cases $\rho_t \leq 4$ and $\rho_t > 4$ in \cref*{eqn:radam-update-rule}. In the ``divergent'' phase, RAdam performs a variant of heavy-ball momentum SGD~\citep{polyak1964momentum}.~\footnote{The departure from standard heavy-ball momentum is in the bias correction by $(1 - \beta_1^t)$.} Then, in the ``convergent'' phase, RAdam performs Adam, with the learning rate scaled down by $\omega_t$.

However, this is \emph{not} dynamic scaling based on the training-time behavior of the optimizer \emph{or} the distribution of the gradients. Rather, $\omega_t$ is a deterministic function of solely $t$ and $\beta_2$. Thus, the ``convergent'' phase is simply Adam with a fixed warmup schedule. We find that for all practically relevant values of $\beta_2$, the condition $\rho_t \leq 4$ is simply $t \leq 4$:

\begin{restatable}[]{fact}{rhotist} \label{fact:rhot4-is-t4}
  Assume that $0.8 \leq \beta_2 < 1$ and $t$ is a positive integer. Then, for $\rho_t$ as defined in \cref*{eqn:radam-rhot}:
  \begin{equation*}
    \rho_t \leq 4 \iff t \leq 4
  \end{equation*}
\end{restatable}
\begin{proof}
See \cref*{apx:misc-derivations-rhot-is-t4}.
\end{proof}

Thus follows a layman's description of RAdam:
\begin{enumerate}
    \item Perform four iterations of heavy-ball momentum.
    \item At iteration five and beyond, use Adam with a fixed warmup schedule.
\end{enumerate}

On its face, using four iterations of momentum at the beginning of training seems arbitrary. In preliminary experimentation (including the experimental settings described in \cref*{sec:results}), we performed ablations over the following options for these four initial iterations:
\begin{itemize}
  \item Do absolutely nothing.
  \item Use Adam with learning rate $\alpha \cdot \omega_5$ (i.e. do exactly what RAdam does at the fifth iteration).
  \item Use Adam with linear warmup to $\alpha \cdot \omega_5$ (i.e. gradually warm up the learning rate to RAdam's fifth iteration).
\end{itemize}
As expected for a decision affecting only four training iterations, the practical difference between these choices is uniformly negligible. Thus, the only possible benefit of RAdam stems from its custom warmup schedule $\omega_t$ for the fifth iteration and beyond. We revisit this topic in \cref*{sec:ruleofthumb,sec:results}.

\subsection{Variance-Based Motivation for RAdam and Warmup}
\label{sec:radam-variance}

Given the arbitrary nature of RAdam's operation, we proceed to investigate the motivation for RAdam, which \citeauthor{liu2019radam} also identify as the underlying motivation for warmup's crucial role in Adam.

\citeauthor{liu2019radam} focus their principal analysis on the term $\sqrt{\frac{1 - \beta_2^t}{v_t}}$. Fixing $\epsilon = 0$, this term can be interpreted as Adam's ``adaptive learning rate'', which scales the global learning rate for each parameter before computing Adam's final update for that parameter. They identify that the quantity $\Var \left[ \sqrt{\frac{1 - \beta_2^t}{v_t}} \right]$ does not exist during the first few training iterations,~\footnote{The authors approximate $\frac{1 - \beta_2^t}{v_t}$ as having a scaled inverse $\chi^2$ distribution, under the assumption that (1) all gradients are i.i.d.\ and zero-mean, and (2) a simple average approximates an exponential moving average.} and even after converging to a finite value, continues to remain elevated for some time.

Perhaps the most immediate observation is that early-stage gradients are \emph{not} zero-mean. In fact, at the beginning of optimization, the expected magnitude of the gradients $\grhL_t(\theta_t)$ (\ie{} absolute value of the deterministic gradients $\grL(\theta_t)$) should dominate the gradient variance, since a randomly-initialized model is exceedingly unlikely to be near a local minimum of $\L(\theta_t)$. Indeed, on a demonstration training run of a feed-forward network on the EMNIST digit recognition task, we observe that the median coefficient of variation of the gradients (\cref*{fig:emnist-median-coeff-variation}) starts at below 1, indicating that for most parameters, the expected value of the gradient exceeds the standard deviation during early-stage training. Only beyond training iteration 50 does the coefficient of variation consistently exceed 1. Relaxing the zero-mean assumption decreases $\Var \left[ \frac{1 - \beta_2^t}{v_t} \right]$ considerably.~\footnote{Although \citeauthor{liu2019radam} do not comment on the relative magnitudes of $\Var \left[ \frac{1 - \beta_2^t}{v_t} \right]$, their Fig. 9 reveals that coefficients of variation below 1 dampen that quantity by an order of magnitude or more.}

More important, however, is that $m_t$ and $v_t$ are not at all independent. \cref*{fig:emnist-moment-corr} reveals that in the EMNIST setting, the absolute value of the first moment estimator ($\abs{m_t}$) is extremely correlated with the square root of the second moment estimator ($\sqrt{v_t}$). Since Adam's parameter updates are proportional to $m_t / \sqrt{v_t}$, high correlation between these two quantities implies that the magnitude of the updates are quite regular, despite the high variance of $\sqrt{\frac{1}{v_t}}$.

Indeed, during the first training iteration ($t = 1$), it is guaranteed that $\abs{m_t} = \sqrt{v_t}$ for all parameters, making all Adam parameter updates either $-\alpha$ or $\alpha$ (assuming $\epsilon = 0$). Thus, even though $\Var \left[ \frac{1 - \beta_2^t}{v_t} \right]$ is divergent, the magnitude of the parameter updates themselves are constant. Ironically, it is \emph{precisely} when the adaptive learning rate's variance is ``divergent'' that the actual parameter update magnitudes have zero variance. This suggests that the adaptive learning rate may not be the best medium of analysis for understanding the role of warmup in Adam.

\subsection{High Initial Update Step Magnitudes Necessitate Warmup in Adam}
\label{sec:radam-update-step-mag}

We provide an alternative view of the frequent necessity of learning rate warmup when using Adam. We do so by directly investigating the magnitudes of the update steps, perhaps the most proximate determinant of training stability.

In stochastic gradient descent, parameter updates are simply the gradients multiplied by the learning rate. Warmup for SGD can thus be motivated as mitigating the large expected magnitudes of the gradients (directly proportional to update magnitudes) and rapid change in gradients at the beginning of training~\citep{goyal2017imagenet, gotmare2019heuristics}. Similar logic can be employed for adaptive methods.

On the other hand, if a model's gradients have near-zero means and low gradient variances, the update steps are similarly well-regulated and optimization via SGD can be stable without any learning rate warmup. For example, a nearly-converged model (thus having near-zero expected gradients and low gradient magnitudes) trained via SGD can have its optimization be stably restarted without learning rate warmup.

This is not the case with Adam. We proceed to computationally analyze the magnitude of Adam's update step over the course of training. Specifically, we demonstrate via simulation that even when the model parameters $\theta_t$ are initialized at an idealized local minimum of $\L(\theta)$ (\ie{} $\grhL_t(\theta_t)$ has zero mean and is i.i.d.\ across time), the magnitude of Adam's update steps will still be quite high at the start of training, only gradually decaying toward a stationary distribution.

\begin{figure}[t]
  \input{fig/simulated_update_distribution.tex}
\end{figure}

\textbf{Simulation configuration} \quad  All gradients are simulated as i.i.d.\, normal variables with zero mean and constant isotropic variance $10^{-9}$, thus approximating the optimization dynamics at an exact local minimum of $\L(\theta)$.~\footnote{Note that the behavior of Adam in this setting is invariant to the choice of variance constant.} We sample independent gradient trajectories (each 1000 iterations long) for 25000 parameters. We then run the Adam optimizer with these sampled gradients and evaluate the distribution of the update step magnitudes (before multiplication by the global learning rate $\alpha$) at each iteration. The Adam optimizer configuration is $\beta_1 = 0.9$, $\beta_2 = 0.999$, and $\epsilon = 0$.

\textbf{Simulation results} \quad \cref*{fig:simulated-update-distribution} depicts the outcome of this computational simulation. As alluded to in \cref*{sec:radam-variance}, the update magnitudes for all parameters start at $1 \cdot \alpha$. The update magnitudes gradually decay but continue to remain high for quite some time, only beginning to settle into a stationary distribution after 40 or so training iterations (with median update magnitude $\approx 0.16 \cdot \alpha$). We extend the trajectory length to 10000 and find that the median update step of the stationary distribution is approximately $0.153 \cdot \alpha$.

These results imply that unlike SGD, Adam will always encounter early-stage training instability by way of large update magnitudes, even when the model is \emph{already initialized at a local minimum}. This stands as a contributing factor to Adam's need for learning rate warmup above and beyond that of first-order methods.

\textbf{Comparison to real-world, random initialization settings}
\quad
Finally, we examine the update step distribution of a model initialized away from a local minimum of $\L(\theta)$. \cref*{fig:emnist-median-update-magnitude} depicts the median parameter update magnitudes of Adam in the EMNIST setting from \cref*{sec:radam-variance}. We observe a qualitative similarity to the local minimum simulation results -- the update magnitudes start at $1 \cdot \alpha$, only gradually settling into a stationary distribution around $0.15 \cdot \alpha$.

Note that the EMNIST optimization decreases more slowly in update magnitude and takes longer ($\approx 100$ training iterations) to settle into the stationary distribution. This suggests that the update step non-regularity observed in the idealized local minimum initialization setting is only exacerbated in the more realistic setting of random initialization.
  \section{Rules of Thumb}
\label{sec:ruleofthumb}

\begin{figure*}[ht]
  \input{fig/effective_warmup_period.tex}
\end{figure*}

Turning to the practical application of learning rate warmup, we first define a simple heuristic function, the \emph{effective warmup period}, to characterize the dampening effect of warmup schedules. We then present and intuitively motivate two Adam warmup schedules that require no tuning and are thus usable as rules of thumb.

\subsection{Effective Warmup Period}

We define the \emph{effective warmup period} $\mathcal{T}(\omega)$ of a warmup schedule $\omega$ as follows:
\begin{equation*}
    \mathcal{T}(\omega) = \sum_{t = 1}^\infty \left( 1 - \omega_t \right)
\end{equation*}
Intuitively, this is the sum of the warmup's dampening effect across all of training.

\subsection{Exponential Warmup}

We propose a simple ``exponential warmup'' schedule based on a decaying exponential and a constant $\tau$:
\begin{equation}
    \omega_t^{\text{expo}, \tau} = 1 - \exp \left( -\frac{1}{\tau} \cdot t \right)
\end{equation}
The constant $\tau$ is analogous to a linear warmup period, and we recommend $\tau = (1 - \beta_2)^{-1}$ as a rule of thumb:
\begin{equation}
    \label{eqn:exp-rule-of-thumb}
    \omega_t^{\text{expo}, \text{untuned}} = 1 - \exp \left( -(1 - \beta_2) \cdot t \right)
\end{equation}

In choosing $\tau$, our guiding (albeit extremely speculative) intuition is to have the warmup factor $\omega_t^{\text{expo}, \tau}$ be roughly equivalent to Adam's second moment bias correction term in Adam. This term, $1 - \beta_2^t$, is the sum of the coefficients in the moving average estimation of the second moment, and can thus be interpreted as how ``complete'' the second moment estimator is at any given point in time. We briefly show the approximate correspondence between the bias correction term and the warmup factor:~\footnote{The second step follows from a first-order Taylor expansion of $\log(\beta_2)$ around $\beta_2 = 1$. In practice, this approximation is extremely accurate for typical values of $\beta_2$.}
\begin{align*}
    1 - \beta_2^t &= 1 - \exp \left( \log(\beta_2) \cdot t \right) \\
    &\approx 1 - \exp \left( (\beta_2 - 1) \cdot t \right) \\
    &= 1 - \exp \left( -(1 - \beta_2) \cdot t \right)
\end{align*}

\subsection{Linear Warmup}

Recall the formulation of linear warmup:
\begin{equation*}
    \omega_t^{\text{linear}, \tau} = \min \left\{ 1, \frac{1}{\tau} \cdot t \right\}
\end{equation*}

As a similar rule of thumb to the exponential warmup schedule, we suggest performing linear warmup over $\tau = 2 \cdot \left( 1 - \beta_2 \right)^{-1}$ iterations:
\begin{equation}
    \label{eqn:linear-rule-of-thumb}
    \omega_t^{\text{linear}, \text{untuned}} = \min \left\{ 1, \frac{1 - \beta_2}{2} \cdot t \right\}
\end{equation}

Our choice of $\tau$ is carried over from exponential warmup as a starting point. To preserve the same effective warmup period, the $\tau$ from the exponential rule-of-thumb is multiplied by 2 to account for the fact that exponential warmup decelerates over time, whereas linear warmup does not. We elaborate in \cref*{apx:misc-derivations-linear-exp-equivalence}.

\subsection{Comparison with RAdam}

We first compare RAdam with the rule-of-thumb schedules (\cref*{eqn:exp-rule-of-thumb,eqn:linear-rule-of-thumb}) by computing their effective warmup periods across a range of $\beta_2$ values.~\footnote{For the purpose of this analysis, $w_{\{1, 2, 3, 4\}}$ are all defined to be zero for RAdam.} \cref*{fig:effective-warmup-period} reveals that the effective warmup periods of RAdam and the rules of thumb are nearly identical across all practical values of $\beta_2$, indicating that they have similar dampening effects over early-stage training.

We then proceed to examine the trajectory of the warmup schedule for the commonly used setting of $\beta_2 = 0.999$. \cref*{fig:warmup-schedule} reveals that the functional forms of the warmup factors are qualitatively similar in magnitudes. The warmup schedules for RAdam and the rule-of-thumb exponential warmup closely correspond in shape as well.

We thus posit that RAdam and the untuned rule-of-thumb warmup schedules are more or less interchangeable. An empirical verification follows.
  \section{Experiments}
\label{sec:results}

We evaluate untuned exponential warmup (\cref*{eqn:exp-rule-of-thumb}), untuned linear warmup (\cref*{eqn:linear-rule-of-thumb}), and RAdam across a variety of supervised machine learning tasks. For brevity, all experimental settings are summarized in the main text and comprehensively detailed in \cref*{apx:training-details}.

\subsection{Image Classification}
\label{sec:results-imagenet}

Using each of the three warmup methods, we train a ResNet-50 model~\citep{he2016resnet} on the ILSVRC (``ImageNet'') image classification dataset with various configurations of Adam. Specifically, we sweep over:
\begin{align*}
    \alpha~\text{(learning rate)} &\in \left\{ {10}^{-4}, {10}^{-3}, {10}^{-2} \right\} \\
    \beta_2 &\in \left\{ 0.99, 0.997, 0.999 \right\}
\end{align*}

\cref*{tbl:summary-imagenet} presents the top-1 error rates at the end of training for the three warmup methods. Across all configurations of Adam, the top-1 error rates are indistinguishable between the warmup methods.~\footnote{The best error rates fall roughly 3\% behind those from SGD, as is typical with Adam on computer vision tasks.}

\begin{table}[ht]
  \centering

\captionsetup{
  justification=centering,
  margin=0.25in
}

\scriptsize

\setlength{\tabcolsep}{3pt}

\begin{tabular}{l l | c c c}
  \toprule
  
  \textbf{LR} & \textbf{$\beta_2$} &          \textbf{Exponential} &               \textbf{Linear} &                \textbf{RAdam} \\
  
  \midrule
  
   $ {10}^{-4} $ &      0.99 &  $ 34.2 \% \pm 0.1 $ &  $ 34.2 \% \pm 0.1 $ &  $ 34.2 \% \pm 0.1 $ \\
 $ {10}^{-4} $ &     0.997 &  $ 34.3 \% \pm 0.2 $ &  $ 34.2 \% \pm 0.2 $ &  $ 34.1 \% \pm 0.1 $ \\
 $ {10}^{-4} $ &     0.999 &  $ 34.5 \% \pm 0.1 $ &  $ 34.4 \% \pm 0.1 $ &  $ 34.2 \% \pm 0.3 $ \\
 $ {10}^{-3} $ &      0.99 &  $ 27.9 \% \pm 0.1 $ &  $ 28.0 \% \pm 0.1 $ &  $ 28.4 \% \pm 0.1 $ \\
 $ {10}^{-3} $ &     0.997 &  $ 27.9 \% \pm 0.1 $ &  $ 27.9 \% \pm 0.1 $ &  $ 28.3 \% \pm 0.1 $ \\
 $ {10}^{-3} $ &     0.999 &  $ 28.2 \% \pm 0.1 $ &  $ 28.3 \% \pm 0.1 $ &  $ 28.4 \% \pm 0.1 $ \\
 $ {10}^{-2} $ &      0.99 &  $ 29.3 \% \pm 0.1 $ &  $ 29.3 \% \pm 0.3 $ &  $ 29.4 \% \pm 0.2 $ \\
 $ {10}^{-2} $ &     0.997 &  $ 29.2 \% \pm 0.2 $ &  $ 29.3 \% \pm 0.1 $ &  $ 29.4 \% \pm 0.5 $ \\
 $ {10}^{-2} $ &     0.999 &  $ 28.9 \% \pm 0.2 $ &  $ 28.7 \% \pm 0.1 $ &  $ 29.8 \% \pm 0.4 $ \\
  
  \bottomrule
\end{tabular}

\caption{Top-1 error rates of ResNet-50 on ImageNet (means and standard deviations over 5 random seeds).}

\label{tbl:summary-imagenet}
\end{table}

\begin{figure}[ht]
  \input{fig/imagenet_train_loss.tex}
\end{figure}

We next examine the course of optimization for individual configurations of Adam's $\alpha$ and $\beta_2$. \cref*{fig:imagenet-train-loss} depicts the training loss using the popular ``default'' Adam configuration of learning rate $\alpha = 10^{-3}$ and $\beta_2 = 0.999$, revealing that the behavior of these warmup methods is indeed nearly indistinguishable.

\cref*{apx:supp-results-imagenet} provides both training and validation metrics (\cref*{fig:imagenet-train-loss-full,fig:imagenet-valid-error-full} respectively) for all tested configurations, reinforcing this trend.

\subsection{Language Modeling}
\label{sec:results-wiki103}

Using each of the three warmup methods, we train a state-of-the-art Transformer-based language model from~\citet{baevski2018adaptivelm} on \texttt{WIKITEXT-103}. We sweep over the following grid of Adam hyperparmeters:
\begin{align*}
    \alpha~\text{(learning rate)} &\in \left\{ 1 \cdot {10}^{-4}, 3 \cdot {10}^{-4}, 5 \cdot {10}^{-4} \right\} \\
    \beta_2 &\in \left\{ 0.99, 0.998, 0.999 \right\}
\end{align*}
with $\beta_1=0.9$ and $\epsilon=10^{-7}$ fixed. As with image classification, we observe in \cref*{tbl:summary-wiki103} that the choice of warmup method has a minimal impact on training across different hyperparameters.

\begin{table}[ht]
  \centering

\captionsetup{
  justification=centering,
  margin=0.25in
}

\scriptsize

\setlength{\tabcolsep}{3pt}

\begin{tabular}{l l | c c c}
  \toprule
  
  \textbf{LR} & \textbf{$\beta_2$} &          \textbf{Exponential} &               \textbf{Linear} &                \textbf{RAdam} \\
  
  \midrule
  
  $1 \cdot {10}^{-4}$ & 0.99& $ 21.0 \pm  0.1 $& $ 21.0 \pm  0.1 $& $ 21.1 \pm  0.1 $\\
$1 \cdot {10}^{-4}$ & 0.998& $ 19.9 \pm  0.0 $& $ 19.9 \pm  0.0 $& $ 20.0 \pm  0.0 $\\
$1 \cdot {10}^{-4}$ & 0.999& $ 20.0 \pm  0.0 $& $ 20.0 \pm  0.0 $& $ 20.1 \pm  0.1 $\\
$3 \cdot {10}^{-4}$ & 0.99& $ 21.3 \pm  0.3 $& $ 20.8 \pm  0.1 $& $ 22.4 \pm  0.0 $\\
$3 \cdot {10}^{-4}$ & 0.998& $ 19.6 \pm  0.0 $& $ 19.6 \pm  0.0 $& $ 19.6 \pm  0.1 $\\
$3 \cdot {10}^{-4}$ & 0.999& $ 19.5 \pm  0.0 $& $ 19.5 \pm  0.0 $& $ 19.5 \pm  0.0 $\\
$5 \cdot {10}^{-4}$ & 0.99& $ 24.4 \pm  2.4 $& $ 24.1 \pm  1.4 $& $ 26.0 \pm  1.8 $\\
$5 \cdot {10}^{-4}$ & 0.998& $ 20.1 \pm  0.0 $& $ 20.0 \pm  0.0 $& $ 20.1 \pm  0.0 $\\
$5 \cdot {10}^{-4}$ & 0.999& $ 19.8 \pm  0.0 $& $ 19.7 \pm  0.1 $& $ 19.7 \pm  0.0 $\\
  
  \bottomrule
\end{tabular}

\caption{Validation perplexity of a Transformer LM on the \texttt{WIKITEXT-103} dataset (means and standard deviations over 3 random seeds).}

\label{tbl:summary-wiki103}
\end{table}

\begin{figure}[ht]
  \input{fig/wiki103_valid_ppl.tex}
\end{figure}

\cref*{fig:wiki103-valid-ppl} depicts the validation perplexity throughout training for the best Adam parametrization ($\alpha=10^{-4}$ and $\beta_2=0.999$), which similarly supports the indistinguishability of the warmup methods.

\subsection{Machine Translation}
\label{sec:results-wmt16-en-de}

Finally, we evaluate the warmup methods on a large scale machine translation task. Using each of the three warmup methods, we train a Transformer model~\citep{vasmani2017transformer} on the WMT16 English-German (``EN-DE'') dataset. We fix Adam's $\beta_1=0.9$ and $\epsilon=10^{-7}$ and sweep over the following grid of Adam hyperparameters:
\begin{align*}
    \alpha~\text{(learning rate)} &\in \left\{ 5 \cdot {10}^{-5}, 8 \cdot {10}^{-5}, 1 \cdot {10}^{-4} \right\} \\
    \beta_2 &\in \left\{ 0.98, 0.99, 0.998, 0.999 \right\}
\end{align*}

\begin{table}[ht]
  \centering

\captionsetup{
  justification=centering,
  margin=0.25in
}

\scriptsize

\setlength{\tabcolsep}{3pt}

\begin{tabular}{l l | c c c}
  \toprule
  
  \textbf{LR} & \textbf{$\beta_2$} &          \textbf{Exponential} &               \textbf{Linear} &                \textbf{RAdam} \\
  
  \midrule
  
  $5 \cdot {10}^{-5}$ & 0.98& $ 24.5 \pm  0.1 $& $ 24.4 \pm  0.1 $& $ 24.4 \pm  0.1 $\\
$5 \cdot {10}^{-5}$ & 0.99& $ 24.5 \pm  0.0 $& $ 24.5 \pm  0.0 $& $ 24.5 \pm  0.1 $\\
$5 \cdot {10}^{-5}$ & 0.998& $ 24.3 \pm  0.2 $& $ 24.4 \pm  0.2 $& $ 24.4 \pm  0.1 $\\
$5 \cdot {10}^{-5}$ & 0.999& $ 24.2 \pm  0.1 $& $ 24.2 \pm  0.1 $& $ 24.1 \pm  0.1 $\\
$8 \cdot {10}^{-5}$ & 0.98& $ 25.9 \pm  0.1 $& $ 25.9 \pm  0.1 $& $ 25.9 \pm  0.1 $\\
$8 \cdot {10}^{-5}$ & 0.99& $ 25.9 \pm  0.2 $& $ 25.9 \pm  0.1 $& $ 25.9 \pm  0.0 $\\
$8 \cdot {10}^{-5}$ & 0.998& $ 26.0 \pm  0.1 $& $ 25.2 \pm  1.0 $& $ 25.9 \pm  0.1 $\\
$8 \cdot {10}^{-5}$ & 0.999& $ 25.7 \pm  0.1 $& $ 25.8 \pm  0.1 $& $ 25.7 \pm  0.0 $\\
$1 \cdot {10}^{-4}$ & 0.98& $ 26.5 \pm  0.1 $& $ 26.6 \pm  0.1 $& $ 26.6 \pm  0.1 $\\
$1 \cdot {10}^{-4}$ & 0.99& $ 26.7 \pm  0.1 $& $ 26.6 \pm  0.1 $& $ 26.6 \pm  0.0 $\\
$1 \cdot {10}^{-4}$ & 0.998& $ 25.9 \pm  0.9 $& $ 26.5 \pm  0.1 $& $ 26.6 \pm  0.0 $\\
$1 \cdot {10}^{-4}$ & 0.999& $ 26.2 \pm  0.2 $& $ 26.4 \pm  0.0 $& $ 26.4 \pm  0.0 $\\
  
  \bottomrule
\end{tabular}

\caption{BLEU score of Transformer on WMT16-EN-DE (means and standard deviations over 3 random seeds).}

\label{tbl:summary-wmt16-en-de}
\end{table}

\begin{figure}[ht]
  \input{fig/wmt16_en_de_valid_ppl.tex}
\end{figure}

We observe no perceptible differences between the warmup methods in either final performance (\cref*{tbl:summary-wmt16-en-de}), or in the training-time metrics of a single canonical configuration ($\alpha=10^{-4}$ and $\beta_2=0.999$, shown in \cref*{fig:wmt16-en-de-valid-ppl}).
  \section{Discussion}
\label{sec:discussion}

We discuss various consequences of our findings, along with directions for future work.

\subsection{Extended Warmup Periods}

The analysis of the update step magnitudes in \cref*{sec:radam-update-step-mag} suggests shorter warmup periods than those typically used in practice. For example, using the setting of $\beta_2 = 0.999$, Adam's update magnitudes in the theoretical model converge to a stationary distribution in roughly 40 iterations. If update magnitudes were the only relevant consideration, then a warmup schedule over a few hundred iterations would suffice to stabilize training. In contrast, the effective warmup periods of both RAdam and our rule-of-thumb schedules are roughly 1000 iterations for $\beta_2 = 0.999$. State-of-the-art methods with hand-tuned warmup schedules often go well beyond, using up to 10000 iterations of linear warmup in some cases~\citep{liu2019roberta,baevski2018adaptivelm,ott2019fairseq}.

Accordingly, we surmise that the precise channel by which Adam necessitates an extended period of warmup is still an unresolved question, likely related to the properties of the gradients at random initialization. Future work could rigorously investigate the effect of extended warmup periods on the training dynamics of Adam, beyond simple per-iteration statistics.

\subsection{Consequences of Update Step Invariance to Gradients}

One ancillary finding of \cref*{sec:radam-update-step-mag} is that the magnitudes of Adam's update steps during later stages of training are largely invariant to the properties or dynamics of the gradient distribution -- both the simulated local optimum and real-world random initialization settings result in convergence to similar stationary distributions of update magnitudes. This suggests that learning rate \emph{decay} at later stages of training could be the only way to improve late-stage convergence, as Adam's late-stage update magnitudes do not appear to be very sensitive to the variance or stationarity of gradients. In particular, we suspect that variance-based methods of improving the late-stage convergence of SGD, such as increasing the batch size~\citep{smith2018batchsize}, will not yield comparable benefits when applied to Adam, as the stationary distribution of the update magnitudes will remain largely the same. Partially adaptable methods~\citep{chen2018padam, keskar2017adamsgd, luo2019adabound}, which interpolate between the full adaptivity of Adam and the non-adaptivity of SGD, may hold more promise for improving late-stage convergence.

\subsection{Dynamic Warmup}

All methods considered by this work use fixed warmup schedules, computed only as a function of the training iteration $t$ and various hyperparameters. Such schedules will inevitably be brittle to some combination of problem setting, model, and optimizer configuration. Another direction for future work could be to devise truly dynamic mechanisms for scheduling warmup in Adam. Such a mechanism could (among other things) track and utilize auxiliary statistics, such as the running moments of the applied updates, in order to determine the stability of training at each iteration.

This direction comes dangerously close to seeking the ``holy grail'' of an automatic learning rate tuner; existing attempts to devise such a method have achieved limited adoption as of yet~\citep{li2017sme, zhang2017yellowfin, baydin2018hgd}. What makes this potentially more tractable is that a maximum learning rate is still tuned and given \emph{a priori} to the optimizer; the task is then restricted to dynamic scheduling of the learning rate from zero to this known constant, instead of an arbitrary range $(0, \infty)$.
  \section{Conclusion}
\label{sec:conclusion}

We show that the Rectified Adam (RAdam) algorithm can be characterized as four steps of momentum SGD, followed by Adam with a fixed warmup schedule. We also examine the shortcomings of a variance-based approach to analyzing the learning rate warmup heuristic, and we illustrate that Adam's frequent need for learning rate warmup can be partially explained by inspecting Adam's early-stage update step magnitudes when applied to an already-converged model.

RAdam's claimed benefits are its superior performance to Adam and its elimination of costly warmup schedule tuning. We obviate RAdam by providing two simple ``rule-of-thumb'' warmup schedules for Adam, both of which require no tuning. Linear warmup of Adam's learning rate over $2 \cdot (1 - \beta_2)^{-1}$ iterations is functionally equivalent to RAdam across a wide range of settings. Hence, we suggest that practitioners considering the need for untuned warmup of Adam's learning rate first try linear warmup over $2 \cdot (1 - \beta_2)^{-1}$ training iterations.

  \newpage
  \bibliography{main}

\begin{thebibliography}{34}
\providecommand{\natexlab}[1]{#1}
\providecommand{\url}[1]{\texttt{#1}}
\providecommand{\urlprefix}{URL }
\expandafter\ifx\csname urlstyle\endcsname\relax
  \providecommand{\doi}[1]{doi:\discretionary{}{}{}#1}\else
  \providecommand{\doi}{doi:\discretionary{}{}{}\begingroup
  \urlstyle{rm}\Url}\fi

\bibitem[{Baevski and Auli(2018)}]{baevski2018adaptivelm}
Baevski, A.; and Auli, M. 2018.
\newblock Adaptive Input Representations for Neural Language Modeling.
\newblock \emph{CoRR} abs/1809.10853.
\newblock \urlprefix\url{http://arxiv.org/abs/1809.10853}.

\bibitem[{Baydin et~al.(2018)Baydin, Cornish, Mart{\'{\i}}nez{-}Rubio, Schmidt,
  and Wood}]{baydin2018hgd}
Baydin, A.~G.; Cornish, R.; Mart{\'{\i}}nez{-}Rubio, D.; Schmidt, M.; and Wood,
  F. 2018.
\newblock Online Learning Rate Adaptation with Hypergradient Descent.
\newblock In \emph{6th International Conference on Learning Representations,
  {ICLR} 2018, Vancouver, BC, Canada, April 30 - May 3, 2018, Conference Track
  Proceedings}.
\newblock \urlprefix\url{https://openreview.net/forum?id=BkrsAzWAb}.

\bibitem[{Brown et~al.(2020)Brown, Mann, Ryder, Subbiah, Kaplan, Dhariwal,
  Neelakantan, Shyam, Sastry, Askell, Agarwal, Herbert{-}Voss, Krueger,
  Henighan, Child, Ramesh, Ziegler, Wu, Winter, Hesse, Chen, Sigler, Litwin,
  Gray, Chess, Clark, Berner, McCandlish, Radford, Sutskever, and
  Amodei}]{brown2020gpt3}
Brown, T.~B.; Mann, B.; Ryder, N.; Subbiah, M.; Kaplan, J.; Dhariwal, P.;
  Neelakantan, A.; Shyam, P.; Sastry, G.; Askell, A.; Agarwal, S.;
  Herbert{-}Voss, A.; Krueger, G.; Henighan, T.; Child, R.; Ramesh, A.;
  Ziegler, D.~M.; Wu, J.; Winter, C.; Hesse, C.; Chen, M.; Sigler, E.; Litwin,
  M.; Gray, S.; Chess, B.; Clark, J.; Berner, C.; McCandlish, S.; Radford, A.;
  Sutskever, I.; and Amodei, D. 2020.
\newblock Language Models are Few-Shot Learners.
\newblock \emph{CoRR} abs/2005.14165.
\newblock \urlprefix\url{https://arxiv.org/abs/2005.14165}.

\bibitem[{Chen and Gu(2018)}]{chen2018padam}
Chen, J.; and Gu, Q. 2018.
\newblock Closing the Generalization Gap of Adaptive Gradient Methods in
  Training Deep Neural Networks.
\newblock \emph{CoRR} abs/1806.06763.
\newblock \urlprefix\url{http://arxiv.org/abs/1806.06763}.

\bibitem[{Cohen et~al.(2017)Cohen, Afshar, Tapson, and van
  Schaik}]{cohen2017emnist}
Cohen, G.; Afshar, S.; Tapson, J.; and van Schaik, A. 2017.
\newblock {EMNIST:} Extending {MNIST} to handwritten letters.
\newblock In \emph{2017 International Joint Conference on Neural Networks,
  {IJCNN} 2017, Anchorage, AK, USA, May 14-19, 2017}, 2921--2926.
\newblock \doi{10.1109/IJCNN.2017.7966217}.
\newblock \urlprefix\url{https://doi.org/10.1109/IJCNN.2017.7966217}.

\bibitem[{Devlin et~al.(2018)Devlin, Chang, Lee, and
  Toutanova}]{devlin2018bert}
Devlin, J.; Chang, M.; Lee, K.; and Toutanova, K. 2018.
\newblock {BERT:} Pre-training of Deep Bidirectional Transformers for Language
  Understanding.
\newblock \emph{CoRR} abs/1810.04805.
\newblock \urlprefix\url{http://arxiv.org/abs/1810.04805}.

\bibitem[{Duchi, Hazan, and Singer(2010)}]{duchi2010adagrad}
Duchi, J.~C.; Hazan, E.; and Singer, Y. 2010.
\newblock Adaptive Subgradient Methods for Online Learning and Stochastic
  Optimization.
\newblock In \emph{{COLT} 2010 - The 23rd Conference on Learning Theory, Haifa,
  Israel, June 27-29, 2010}, 257--269.
\newblock
  \urlprefix\url{http://colt2010.haifa.il.ibm.com/papers/COLT2010proceedings.pdf\#page=265}.

\bibitem[{Gehring et~al.(2017)Gehring, Auli, Grangier, Yarats, and
  Dauphin}]{gehring2017convs2s}
Gehring, J.; Auli, M.; Grangier, D.; Yarats, D.; and Dauphin, Y.~N. 2017.
\newblock Convolutional Sequence to Sequence Learning.
\newblock In \emph{Proceedings of the 34th International Conference on Machine
  Learning, {ICML} 2017, Sydney, NSW, Australia, 6-11 August 2017}, 1243--1252.
\newblock \urlprefix\url{http://proceedings.mlr.press/v70/gehring17a.html}.

\bibitem[{Gotmare et~al.(2019)Gotmare, Keskar, Xiong, and
  Socher}]{gotmare2019heuristics}
Gotmare, A.; Keskar, N.~S.; Xiong, C.; and Socher, R. 2019.
\newblock A Closer Look at Deep Learning Heuristics: Learning rate restarts,
  Warmup and Distillation.
\newblock In \emph{7th International Conference on Learning Representations,
  {ICLR} 2019, New Orleans, LA, USA, May 6-9, 2019}.
\newblock \urlprefix\url{https://openreview.net/forum?id=r14EOsCqKX}.

\bibitem[{Goyal et~al.(2017)Goyal, Doll{\'{a}}r, Girshick, Noordhuis,
  Wesolowski, Kyrola, Tulloch, Jia, and He}]{goyal2017imagenet}
Goyal, P.; Doll{\'{a}}r, P.; Girshick, R.~B.; Noordhuis, P.; Wesolowski, L.;
  Kyrola, A.; Tulloch, A.; Jia, Y.; and He, K. 2017.
\newblock Accurate, Large Minibatch {SGD}: Training {ImageNet} in 1 Hour.
\newblock \emph{CoRR} abs/1706.02677.
\newblock \urlprefix\url{http://arxiv.org/abs/1706.02677}.

\bibitem[{He et~al.(2016)He, Zhang, Ren, and Sun}]{he2016resnet}
He, K.; Zhang, X.; Ren, S.; and Sun, J. 2016.
\newblock Deep Residual Learning for Image Recognition.
\newblock In \emph{2016 {IEEE} Conference on Computer Vision and Pattern
  Recognition, {CVPR} 2016, Las Vegas, NV, USA, June 27-30, 2016}, 770--778.
\newblock \doi{10.1109/CVPR.2016.90}.
\newblock \urlprefix\url{https://doi.org/10.1109/CVPR.2016.90}.

\bibitem[{Hinton, Srivastava, and Swersky(2012)}]{hinton2012rmsprop}
Hinton, G.; Srivastava, N.; and Swersky, K. 2012.
\newblock Neural networks for machine learning: Lecture 6a.

\bibitem[{Keskar and Socher(2017)}]{keskar2017adamsgd}
Keskar, N.~S.; and Socher, R. 2017.
\newblock Improving Generalization Performance by Switching from Adam to {SGD}.
\newblock \emph{CoRR} abs/1712.07628.
\newblock \urlprefix\url{http://arxiv.org/abs/1712.07628}.

\bibitem[{Kingma and Ba(2014)}]{kingma2014adam}
Kingma, D.~P.; and Ba, J. 2014.
\newblock Adam: {A} Method for Stochastic Optimization.
\newblock \emph{CoRR} abs/1412.6980.
\newblock \urlprefix\url{http://arxiv.org/abs/1412.6980}.

\bibitem[{Krizhevsky, Sutskever, and Hinton(2012)}]{krizhevsky2012alexnet}
Krizhevsky, A.; Sutskever, I.; and Hinton, G.~E. 2012.
\newblock ImageNet Classification with Deep Convolutional Neural Networks.
\newblock In \emph{Advances in Neural Information Processing Systems 25: 26th
  Annual Conference on Neural Information Processing Systems 2012. Proceedings
  of a meeting held December 3-6, 2012, Lake Tahoe, Nevada, United States.},
  1106--1114.
\newblock
  \urlprefix\url{http://papers.nips.cc/paper/4824-imagenet-classification-with-deep-convolutional-neural-networks}.

\bibitem[{Li, Tai, and E(2017)}]{li2017sme}
Li, Q.; Tai, C.; and E, W. 2017.
\newblock Stochastic Modified Equations and Adaptive Stochastic Gradient
  Algorithms.
\newblock In \emph{Proceedings of the 34th International Conference on Machine
  Learning, {ICML} 2017, Sydney, NSW, Australia, 6-11 August 2017}, 2101--2110.
\newblock \urlprefix\url{http://proceedings.mlr.press/v70/li17f.html}.

\bibitem[{Liu et~al.(2020)Liu, Jiang, He, Chen, Liu, Gao, and
  Han}]{liu2019radam}
Liu, L.; Jiang, H.; He, P.; Chen, W.; Liu, X.; Gao, J.; and Han, J. 2020.
\newblock On the Variance of the Adaptive Learning Rate and Beyond.
\newblock In \emph{International Conference on Learning Representations}.
\newblock \urlprefix\url{https://openreview.net/forum?id=rkgz2aEKDr}.

\bibitem[{Liu et~al.(2019{\natexlab{a}})Liu, Albanie, Nagrani, and
  Zisserman}]{liu2019video}
Liu, Y.; Albanie, S.; Nagrani, A.; and Zisserman, A. 2019{\natexlab{a}}.
\newblock Use What You Have: Video retrieval using representations from
  collaborative experts.
\newblock In \emph{30th British Machine Vision Conference 2019, {BMVC} 2019,
  Cardiff, UK, September 9-12, 2019}, 279. {BMVA} Press.
\newblock
  \urlprefix\url{https://bmvc2019.org/wp-content/uploads/papers/0363-paper.pdf}.

\bibitem[{Liu et~al.(2019{\natexlab{b}})Liu, Ott, Goyal, Du, Joshi, Chen, Levy,
  Lewis, Zettlemoyer, and Stoyanov}]{liu2019roberta}
Liu, Y.; Ott, M.; Goyal, N.; Du, J.; Joshi, M.; Chen, D.; Levy, O.; Lewis, M.;
  Zettlemoyer, L.; and Stoyanov, V. 2019{\natexlab{b}}.
\newblock RoBERTa: {A} Robustly Optimized {BERT} Pretraining Approach.
\newblock \emph{CoRR} abs/1907.11692.
\newblock \urlprefix\url{http://arxiv.org/abs/1907.11692}.

\bibitem[{Loshchilov and Hutter(2019)}]{loshchilov2019adamw}
Loshchilov, I.; and Hutter, F. 2019.
\newblock Decoupled Weight Decay Regularization.
\newblock In \emph{7th International Conference on Learning Representations,
  {ICLR} 2019, New Orleans, LA, USA, May 6-9, 2019}.
\newblock \urlprefix\url{https://openreview.net/forum?id=Bkg6RiCqY7}.

\bibitem[{Luo et~al.(2019)Luo, Xiong, Liu, and Sun}]{luo2019adabound}
Luo, L.; Xiong, Y.; Liu, Y.; and Sun, X. 2019.
\newblock Adaptive Gradient Methods with Dynamic Bound of Learning Rate.
\newblock In \emph{7th International Conference on Learning Representations,
  {ICLR} 2019, New Orleans, LA, USA, May 6-9, 2019}.
\newblock \urlprefix\url{https://openreview.net/forum?id=Bkg3g2R9FX}.

\bibitem[{Ma and Yarats(2019)}]{ma2019qh}
Ma, J.; and Yarats, D. 2019.
\newblock Quasi-hyperbolic momentum and Adam for deep learning.
\newblock In \emph{7th International Conference on Learning Representations,
  {ICLR} 2019, New Orleans, LA, USA, May 6-9, 2019}.
\newblock \urlprefix\url{https://openreview.net/forum?id=S1fUpoR5FQ}.

\bibitem[{Nesterov(1983)}]{nesterov1983nag}
Nesterov, Y.~E. 1983.
\newblock A method for solving the convex programming problem with convergence
  rate $\mathcal{O}(1/k^2)$.
\newblock In \emph{Dokl. Akad. Nauk SSSR}, volume 269, 543--547.

\bibitem[{Nguyen and Salazar(2019)}]{nguyen2019trans}
Nguyen, T.~Q.; and Salazar, J. 2019.
\newblock Transformers without Tears: Improving the Normalization of
  Self-Attention.
\newblock \emph{CoRR} abs/1910.05895.
\newblock \urlprefix\url{http://arxiv.org/abs/1910.05895}.

\bibitem[{Ott et~al.(2019)Ott, Edunov, Baevski, Fan, Gross, Ng, Grangier, and
  Auli}]{ott2019fairseq}
Ott, M.; Edunov, S.; Baevski, A.; Fan, A.; Gross, S.; Ng, N.; Grangier, D.; and
  Auli, M. 2019.
\newblock fairseq: A Fast, Extensible Toolkit for Sequence Modeling.
\newblock In \emph{Proceedings of NAACL-HLT 2019: Demonstrations}.

\bibitem[{Paszke et~al.(2016)Paszke, Gross, Chintala, Chanan, Yang, DeVito,
  Lin, Desmaison, Antiga, and Lerer}]{paszke2016pytorchexamples}
Paszke, A.; Gross, S.; Chintala, S.; Chanan, G.; Yang, E.; DeVito, Z.; Lin, Z.;
  Desmaison, A.; Antiga, L.; and Lerer, A. 2016.
\newblock {PyTorch} Examples.
\newblock \url{https://github.com/pytorch/examples}.

\bibitem[{Paszke et~al.(2017)Paszke, Gross, Chintala, Chanan, Yang, DeVito,
  Lin, Desmaison, Antiga, and Lerer}]{paszke2017pytorch}
Paszke, A.; Gross, S.; Chintala, S.; Chanan, G.; Yang, E.; DeVito, Z.; Lin, Z.;
  Desmaison, A.; Antiga, L.; and Lerer, A. 2017.
\newblock Automatic differentiation in {PyTorch}.
\newblock In \emph{NIPS-W}.

\bibitem[{Polyak(1964)}]{polyak1964momentum}
Polyak, B. 1964.
\newblock Some methods of speeding up the convergence of iteration methods.
\newblock \emph{USSR Computational Mathematics and Mathematical Physics} 4(5):
  1--17.

\bibitem[{Radford et~al.(2019)Radford, Wu, Child, Luan, Amodei, and
  Sutskever}]{radford2019gpt2}
Radford, A.; Wu, J.; Child, R.; Luan, D.; Amodei, D.; and Sutskever, I. 2019.
\newblock Language models are unsupervised multitask learners.
\newblock \emph{{OpenAI Blog}} .

\bibitem[{Russakovsky et~al.(2015)Russakovsky, Deng, Su, Krause, Satheesh, Ma,
  Huang, Karpathy, Khosla, Bernstein, Berg, and
  Fei-Fei}]{russakovsky2015imagenet}
Russakovsky, O.; Deng, J.; Su, H.; Krause, J.; Satheesh, S.; Ma, S.; Huang, Z.;
  Karpathy, A.; Khosla, A.; Bernstein, M.; Berg, A.~C.; and Fei-Fei, L. 2015.
\newblock {ImageNet Large Scale Visual Recognition Challenge}.
\newblock \emph{International Journal of Computer Vision (IJCV)} 115(3):
  211--252.
\newblock \doi{10.1007/s11263-015-0816-y}.

\bibitem[{Smith et~al.(2018)Smith, Kindermans, Ying, and
  Le}]{smith2018batchsize}
Smith, S.~L.; Kindermans, P.; Ying, C.; and Le, Q.~V. 2018.
\newblock Don't Decay the Learning Rate, Increase the Batch Size.
\newblock In \emph{6th International Conference on Learning Representations,
  {ICLR} 2018, Vancouver, BC, Canada, April 30 - May 3, 2018, Conference Track
  Proceedings}.
\newblock \urlprefix\url{https://openreview.net/forum?id=B1Yy1BxCZ}.

\bibitem[{Vaswani et~al.(2017)Vaswani, Shazeer, Parmar, Uszkoreit, Jones,
  Gomez, Kaiser, and Polosukhin}]{vasmani2017transformer}
Vaswani, A.; Shazeer, N.; Parmar, N.; Uszkoreit, J.; Jones, L.; Gomez, A.~N.;
  Kaiser, L.; and Polosukhin, I. 2017.
\newblock Attention Is All You Need.
\newblock \emph{CoRR} abs/1706.03762.
\newblock \urlprefix\url{http://arxiv.org/abs/1706.03762}.

\bibitem[{Yamamoto, Song, and Kim(2020)}]{yamamoto2020parallel}
Yamamoto, R.; Song, E.; and Kim, J.-M. 2020.
\newblock Parallel WaveGAN: A fast waveform generation model based on
  generative adversarial networks with multi-resolution spectrogram.
\newblock In \emph{ICASSP 2020-2020 IEEE International Conference on Acoustics,
  Speech and Signal Processing (ICASSP)}, 6199--6203. IEEE.

\bibitem[{Zhang, Mitliagkas, and R{\'{e}}(2017)}]{zhang2017yellowfin}
Zhang, J.; Mitliagkas, I.; and R{\'{e}}, C. 2017.
\newblock YellowFin and the Art of Momentum Tuning.
\newblock \emph{CoRR} abs/1706.03471.
\newblock \urlprefix\url{http://arxiv.org/abs/1706.03471}.

\end{thebibliography}
  
  \onecolumn
  \appendix
  
  \newpage
  \section{Full details of experimental setup}
\label{apx:training-details}

\subsection{System configuration}

All experiments are performed using Python 3.7 and PyTorch version 1.2~\citep{paszke2017pytorch} compiled with CUDA 10, on Ubuntu 18.04 systems containing 8 NVIDIA V100 GPUs each.

\subsection{Image classification}
\label{apx:training-details-imagenet}

Experimentation is performed using the ILSVRC 2012 1000-class dataset~\citep[``ImageNet'';][]{russakovsky2015imagenet} and a 50-layer convolutional residual network model~\citep[``ResNet-50'';][]{he2016resnet}. The implementation follows that of \citet{paszke2016pytorchexamples},~\footnote{Commit hash \texttt{ee964a2}.} with the only deviations being to enable alternative optimizer configurations, to enable intermediate metric logging, and to drop the last batch from each training epoch.

Training occurs over 90 epochs, with ten-fold learning rate decay after epochs 30 and 60. The minibatch size is 1024. The optimization objective is cross-entropy, with a decoupled weight decay~\citep{loshchilov2019adamw} of $10^{-4}$.

Data augmentation includes horizontal flipping at random, as well as random 224-pixel crops. Validation is performed on 224-pixel center crops.

For Adam and RAdam, the following hyperparameters are fixed: $\beta_1 = 0.9$ and $\eps = 10^{-8}$. All other Adam parameters (warmup schedule, learning rate $\alpha$, and $\beta_2$) are enumerated via parameter sweep as described in \cref*{sec:results-imagenet}. Each Adam configuration is independently trained with 5 random seeds.

\subsection{Language modeling}
\label{apx:training-details-wiki103}

We evaluate the state-of-the-art, Transformer-based language model described in~\cite{baevski2018adaptivelm} on the \texttt{WIKITEXT-103} dataset, consisting of 100M tokens with a size-260K vocabulary. We leverage the author's implementation provided in fairseq~\citep{gehring2017convs2s, ott2019fairseq}, and train on 8 GPUs with half-precision floating point.

Our experimentation setup closely follows~\cite{baevski2018adaptivelm}, except that we sweep over Adam parameters, such as warmup schedule, learning rate $\alpha$, and $\beta_2$, while keeping $\beta_1=0.9$ and $\eps=10^{-7}$ fixed (both for Adam and RAdam). The hyperparameter grid is presented in~\cref*{sec:results-wiki103}. Each Adam configuration is independently trained with 3 random seeds.

\subsection{Machine translation}
\label{apx:training-details-wmt16-en-de}

Our setup employs a state-of-the-art Transformer model~\citep{vasmani2017transformer} implemented in fairseq~\citep{ott2019fairseq}. We train on the WMT16 English-German large machine translation dataset, and evaluate on the \textit{newstest14} validation set.

As observed in~\cite{ma2019qh}, these state-of-the-art large-scale models are fragile to train with Adam and require either a carefully chosen optimization procedure, or robust optimizers that can sustain gradients with large variance, such as QHAdam~\citep{ma2019qh}. To eliminate this factor from our studies, we choose to lower the learning rate $\alpha$ to stabilize training, taking a marginal performance hit in training. Apart from that, our experimentation setup is identical to the one in~\cite{ott2019fairseq}.

We fix Adam parameters $\beta_1=0.9$ and $\epsilon=10^{-7}$, and sweep over the warmup schedule, learning rate $\alpha$, and $\beta_2$, as described in~\cref*{sec:results-wmt16-en-de}. We again use half-precision floating point and train on 8 GPUs. As~\cite{ott2019fairseq} trains on 128 GPUs, we accumulate gradients over $16$ minibatches before each optimization step to achieve an identical configuration. The BLEU score is averaged over $3$ random seeds.

\subsection{Gradient analysis workhorse: EMNIST digit classification}
\label{apx:training-details-emnist}

The EMNIST digit classification task~\citep{cohen2017emnist} serves as the workhorse for our gradient analysis studies. Our model is a simple feed-forward neural network with three hidden layers (sizes 200, 100, and 50) and uniform weight initialization with range inversely proportional to the square root of layer sizes.

Optimization is performed on the cross-entropy objective with the Adam optimizer. The Adam configuration is $\alpha = 10^{-3}$, $\beta_1 = 0.9$, $\beta_2 = 0.999$, $\eps = 10^{-8}$, and decoupled weight decay $10^{-4}$. The minibatch size is 256,

Training occurs over 10000 training iterations. At each training iteration, 256 backwards passes are performed with independently sampled batches to collect a sample of the gradient distribution. Due to the cost of storing and analyzing the gradients of all parameters, we randomly sample 500 parameters from each weight matrix and only collect gradients for the sampled parameters. These samples are used to approximate the distribution of the gradient coefficients of variation. After the 256 backwards passes, one final pass is performed as a regular optimization step to update the model parameters and proceed to the next iteration.
  \newpage
  \section{Miscellaneous derivations}
\label{apx:misc-derivations}

This appendix provides miscellaneous informal derivations of statements in the main text.

\subsection{Number of RAdam momentum iterations}
\label{apx:misc-derivations-rhot-is-t4}

\rhotist*

\begin{proof}

We define $\rho(t, \beta_2)$ to be the continuous version of $\rho_t$, parameterized over both $t$ and $\beta_2$:
\begin{equation*}
    \rho(t, \beta_2) = \frac{2}{1 - \beta_2} - 1 - \frac{2 \cdot t \cdot \beta_2^t}{1 - \beta_2^t}
\end{equation*}

We then differentiate with respect to $t$:
\begin{align*}
    \frac{\partial \rho(t, \beta_2)}{\partial t} &= 2 \cdot \beta_2^t \cdot \frac{\beta_2^t - 1 - t \cdot \ln \beta_2}{\left( 1 - \beta_2^t \right)^2}
\end{align*}
$\frac{\partial \rho(t, \beta_2)}{\partial t}$ is thus positive for all $t > 0$.

We also differentiate with respect to $\beta_2$, and take specific values thereof:
\begin{align*}
    \frac{\partial \rho(t, \beta_2)}{\partial \beta_2} &= 2 \cdot \left( \frac{1}{(1 - \beta_2)^2} - \frac{t^2 \cdot \beta_2^{t - 1}}{(1 - \beta_2^t)^2} \right) \\
    \frac{\partial \rho(4, \beta_2)}{\partial \beta_2} &= 2 \cdot \left( \frac{1}{(1 - \beta_2)^2} - \frac{16 \cdot \beta^3}{(1 - \beta_2^4)^2} \right) \\
    \frac{\partial \rho(5, \beta_2)}{\partial \beta_2} &= 2 \cdot \left( \frac{1}{(1 - \beta_2)^2} - \frac{25 \cdot \beta_2^4}{(1 - \beta_2^5)^2} \right)
\end{align*}
$\frac{\partial \rho(t, \beta_2)}{\partial \beta_2}$ is thus positive for all $\beta_2 \in (0, 1)$ at $t = 4$ and $t = 5$.

Then, we take $\lim\limits_{\beta_2 \rightarrow 1} \rho(4, \beta_2)$:
\begin{align*}
    \lim\limits_{\beta_2 \rightarrow 1} \rho(4, \beta_2) &= \lim\limits_{\beta_2 \rightarrow 1} \left( \frac{2}{1 - \beta_2} - \frac{8 \cdot \beta_2^4}{1 - \beta_2^4} \right) - 1 \\
    &= \lim\limits_{\beta_2 \rightarrow 1} \left( \frac{2 \cdot \left( 4 \cdot \beta_2^3 + 3 \cdot \beta_2^2 + 2 \cdot \beta_2 + 1 \right)}{(1 + \beta_2) (1 + \beta_2^2)} \right) - 1 \\
    &= 5 - 1 \\
    &= 4
\end{align*}

Combining this result with the fact that $\frac{\partial \rho(4, \beta_2)}{\partial \beta_2}$ is positive for $\beta_2 \in (0, 1)$, it follows that $\rho(4, \beta_2) < 4$ for all $\beta_2 \in (0, 1)$. Then, since $\frac{\partial \rho(t, \beta_2)}{\partial t} > 0$ for all $t > 0$, we have that $\rho(t, \beta_2) < 4$ for all $\beta_2 \in (0, 1)$ and $t \in (0, 4]$. We have thus shown that $t \leq 4 \implies \rho_t \leq 4$ for positive integers $t$.

In the reverse direction, we evaluate $\rho(5, 0.8)$:
\begin{align*}
    \rho(5, 0.8) &= \frac{2}{1 - 0.8} - 1 - \frac{2 \cdot 5 \cdot {0.8}^5}{1 - {0.8}^5} \\
    &\approx 9 - 4.87 \\
    &\approx 4.14
\end{align*}

Similarly combining this result with the fact that $\frac{\partial \rho(5, \beta_2)}{\partial \beta_2}$ is positive for $\beta_2 \in (0, 1)$, then with the fact that $\frac{\partial \rho(t, \beta_2)}{\partial t} > 0$ for all $t > 0$, we have that $\rho(t, \beta_2) \gtrsim 4.14$ for all $t \geq 5, \beta_2 \in [0.8, 1)$. We have thus shown that $t > 4 \implies \rho_t > 4$ for positive integers $t$, completing the proof.

\end{proof}

\subsection{Linear warmup period (rule-of-thumb)}
\label{apx:misc-derivations-linear-exp-equivalence}

We desire for the effective warmup period to be roughly equivalent between the exponential and linear rule-of-thumb schedules -- that is, $\mathcal{T}(w^{\text{expo}, \text{untuned}}) \approx \mathcal{T}(w^{\text{linear}, \tau})$. Solving approximately for $\tau$:
\begin{align*}
    \mathcal{T}(w^{\text{expo}, \text{untuned}}) &= \sum_{t = 1}^\infty \exp \left( - (1 - \beta_2) \cdot t \right) \\
    &= \frac{1}{\exp \left( 1 - \beta_2 \right) - 1} \\
    & \approx (1 - \beta_2)^{-1} \\
    \mathcal{T}(w^{\text{linear}, \tau}) &= \sum_{t = 1}^{\tau} \left[ 1 - \frac{1}{\tau} \cdot t \right] = \frac{\tau - 1}{2} \\
    &\approx \frac{\tau}{2} \\
    \tau = 2 \cdot \left(1 - \beta_2\right)^{-1} &\implies \mathcal{T}(w^{\text{expo}, \text{untuned}}) \approx \mathcal{T}(w^{\text{linear}, \tau})
\end{align*}
  \newpage
  \section{Supplementary experimental results}
\label{apx:supp-results}

\subsection{Image classification}
\label{apx:supp-results-imagenet}

\begin{figure}[!th]
  \input{fig/imagenet_train_loss_full.tex}
\end{figure}
\begin{figure}[!th]
  \input{fig/imagenet_valid_error_full.tex}
\end{figure}
}

\end{document}